\documentclass[conference]{IEEEtran}
\IEEEoverridecommandlockouts
\usepackage{cite}
\usepackage{amsmath,amssymb,amsfonts}
\usepackage{algorithmic}
\usepackage{graphicx}
\usepackage{textcomp}
\usepackage{xcolor}
\def\BibTeX{{\rm B\kern-.05em{\sc i\kern-.025em b}\kern-.08em
    T\kern-.1667em\lower.7ex\hbox{E}\kern-.125emX}}

\usepackage{xspace}
\usepackage[size=tiny]{todonotes}
\usepackage{booktabs}
\usepackage{enumitem}
\usepackage{multicol} 
\usepackage{multirow}
\usepackage{tabularx}
\usepackage{arydshln} 
\usepackage{amsfonts}
\usepackage{amsmath}
\usepackage{amssymb}
\usepackage[hidelinks]{hyperref}
\usepackage{url}

\newcommand{\corpus}{\textsc{DeuPARL}\xspace}
\usepackage{subcaption}

\begin{document}

\title{Diachronic Analysis of German Parliamentary Proceedings: 
Ideological Shifts through the Lens of Political Biases
}

\author{\IEEEauthorblockN{Tobias Walter,\textsuperscript{1} Celina Kirschner,\textsuperscript{1} Steffen Eger,\textsuperscript{2} Goran Glava\v{s},\textsuperscript{1} Anne Lauscher,\textsuperscript{1} Simone Paolo Ponzetto\textsuperscript{1}}
\IEEEauthorblockA{
\textsuperscript{1}\textit{Data and Web Science Group}, \textit{University of Mannheim}, Mannheim, Germany\\ 
\textsuperscript{2}\textit{Natural Language Learning Group}, \textit{Technische Universität  Darmstadt}, Darmstadt, Germany\\
celina.kirschner@hotmail.de, tobias.walter-ul@web.de,\\ \{goran, anne, simone\}@informatik.uni-mannheim.de, eger@aiphes.tu-darmstadt.de}}

\maketitle

\begin{abstract}
We analyze bias in historical corpora as encoded in diachronic distributional semantic models by focusing on two specific forms of bias, namely a political (i.e., anti-communism) and racist (i.e., antisemitism) one. For this, we use a new corpus of German parliamentary proceedings, \corpus, spanning the period 1867--2020. We complement this analysis of historical biases in diachronic word embeddings with a novel measure of bias on the basis of term co-occurrences and graph-based label propagation. The results of our bias measurements align with commonly perceived historical trends of antisemitic and anti-communist biases in German politics in different time periods, thus indicating the viability of analyzing historical bias trends using semantic spaces induced from historical corpora.
\end{abstract}

\section{Introduction}
Recent years have seen much work on the topic of bias in data-driven Artificial Intelligence \cite{ntoutsi2020bias} and Machine Learning \cite{Mehrabi19}. In the case of Natural Language Processing, researchers have investigated the bias encoded within semantic spaces 
induced 
by both non-contextualized \cite{caliskan2017semantics,Garg.2018} and contextualized embeddings \cite{kurita-etal-2019-measuring}, and a variety of methods has been developed to debias such embedding spaces in such a way as to make them fairer \cite[\emph{inter alia}]{bolukbasi2016man,gonen2019lipstick,lauscher2020general}.

However, despite much interest on studying and mitigating bias in semantic spaces, few works -- \cite{Garg.2018, Tripodi.} being notable exceptions -- have taken a historical perspective and looked at ways to study bias in diachronic corpora through the lens of distributional embedding models. In this paper, we create a new historical corpus of parliamentary proceedings spanning three different centuries and set to quantify different kinds of bias 
in 
the underlying historical periods. 
Our \corpus corpus covers the protocols of both the German Reichstag and the Bundestag from 1867 until 2020, thus spanning a large timeline that covers many crucial modern and contemporary events (e.g., two world wars): 
it consists of the digitized scans of the older German Reichstag and the digitally-born newer German Bundestag parliamentary proceedings. 
In this corpus, we focus on 
political and racist forms of bias, i.e., anti-communism and antisemitism, respectively. 
We achieve this by building upon previous work on quantifying the biases  
in textual corpora, that is: (1) we devise a number of term-based bias specifications that provide us with an operational definition of antisemitic and anti-communist bias; (2) we quantify the degree of such biases using measures applied to word embeddings induced from different time slices of our corpus; (3) we complement this study of bias in dense semantic vectors with an analysis of bias in more `classic' (i.e., sparse) distributional word representations using a label propagation algorithm applied to word co-occurrence graphs.  As a result, we are able to provide a diachronic analysis of the bias trends in our historical corpus for different kinds of biases and word representations on the basis of our specifications.

\vspace{1em}
The contributions of this work are the following ones:

\begin{itemize}
    \item We present a new historical corpus of parliamentary proceedings built from the protocols of both German Reichstag and Bundestag between 1867 and 2020.
    \item We induce a variety of distributional semantic spaces, both dense and sparse, from different slices of our historical corpus and observe bias trends across time for different kinds of bias (i.e., antisemitism and anti-communism).
    \item We introduce a new measure of bias on the basis of term co-occurrences and graph-based label propagation, which makes it possible to quantify bias in sparse distributional spaces and thus in an arguably more interpretable way.
\end{itemize}

Our results are consistent with common historical interpretation of development of bias in German history, i.e., we identify an increase in antisemitism towards the NS period, with a peak in the Weimar Republic. 
This indicates 
the viability of studying the evolution of bias from a historical perspective on the basis of word embeddings, on associated tests, from historical corpora. We make all code and data (including \corpus) available at \url{https://github.com/umanlp/crosstemporal_bias}.

\section{\corpus: German Parliamentary Corpus}
\label{sec:data}
Our new historical corpus consisting of German parliamentary proceedings from 1867 until 2020, which we refer to as \corpus,\footnote{We make \corpus available at: \url{https://tudatalib.ulb.tu-darmstadt.de/handle/tudatalib/2889}} is built from two main sources.

\begin{table*}[!htb]
    \caption{Excerpt of a speech in a parliamentary session, from 1914. Left: OCR-scanned document. Right: Corrected version. Words containing OCR errors are underlined.}
    \centering
    \begin{tabular}{l|l}
    \toprule
         Gröber, Abgeordneter: Meine Herren, der Gedanke & Gröber, Abgeordneter: Meine Herren, der Gedanke\\
des Herrn Abgeordneten \underline{Gothetn}, der dem Beschluß \underline{deS}
&
des Herrn Abgeordneten Gothein, der dem Beschluß des
\\
hohen Hauses in der letzten Sitzung zu Grunde lag, ist &
hohen Hauses in der letzten Sitzung zu Grunde lag, ist
\\
meines Erachtens zweifellos richtig. Es ist nicht er-
&
meines Erachtens zweifellos richtig. Es ist nicht er-
\\
\bottomrule
    \end{tabular}
    \label{table:ocrerrors}
\end{table*}

\vspace{1em}
\noindent
\textbf{Reichstagsprotokolle.} For parliamentary speeches before 1945, we use the German \emph{Reichstagsprotokolle} 
which are accessible at  \url{https://www.reichstagsprotokolle.de/}. 
The \emph{Bayerische Staatsbibliothek} distributes the digitized data upon request, arranged 
in three splits covering the years 1867--1895 (Norddeutscher Bund / Zollparlamente), 1895--1918 (Kaiserreich), and  
1918--1942 (Weimarer Republik / Nationalsozialismus), respectively. 
These were OCR-scanned using the Abbyy FineReader software.\footnote{\url{https://pdf.abbyy.com/}} To check data quality, we extracted a sample of 
several thousand 
lines which we corrected by hand. In this sample, 
about 23\% of all lines contained one or more OCR errors. 
A sample of the data is shown in Table \ref{table:ocrerrors}: we found the data to be of sufficiently high quality for our purposes, not urgently necessitating further OCR post-correction steps \cite{EgerBrueckMehler2017}. 
Overall, we extracted 5,446 individual protocols in the Reichstagsprotokolle, i.e.,\ stenographic documents corresponding to the parliamentary sessions.

\vspace{1em}
\noindent
\textbf{Bundestagsprotokolle.} For parliamentary speeches after 1945, we use the \emph{Bundestagsprotokolle} from 
\url{https://www.bundestag.de/protokolle}.\footnote{Note that, for the time period of the Division of Germany between 1949 and 1990, the Bundestagsprotokolle only cover the protocols of the Federal Republic of Germany (also known as ``West Germany'').}
We extracted 4,260 
protocols 
in 19 legislative periods from 1949 to March 2020. 

\vspace{1em}
\noindent
\textbf{Preprocessing.} We conducted 
preprocessing steps for \corpus (mostly affecting the Reichstagsprotokolle) including:\footnote{We also aimed at historical spelling normalization, but found it to lead to substantial amount of undesired text modifications.}

\begin{itemize}[leftmargin=4mm]
    \item We fixed word segmentations at line breaks \\
    (e.g., ``Poli- tik'' $\rightarrow$ ``Politik''); 
    \item We lowercased and lemmatized all words (the latter using GermaLemma++ \cite{ortmann2019evaluating}) to increase statistical support for all subsequent models built on top of our data;
    \item We resolved cases such as ``Luft- und Raumfahrt'' into \\ ``Luftfahrt und Raumfahrt'' and fixed erroneous word segmentations (e.g., where numbers and words were merged)
    \item 
    We took care to keep only text from the individual parliamentary sessions, not considering various attachments to the protocols such as lists of names. To this end, we used regular expressions to extract the text within boundaries such as ``Die Sitzung ist eröffnet'' (``\emph{the session is openend}'') and ``Schluss der Sitzung \emph{hh} Uhr \emph{mm} Minute(n)'' (``\emph{end of session hh:mm}'')
\end{itemize}

\vspace{1em}
\noindent
\textbf{Temporal splits.} We defined our own temporal splits for a historical analysis of bias, taking 
historically accepted time periods in modern German history as reference. These are: the Kaiserreich I (KR1, 1867--1890), Kaiserreich II (KR2, 1890--1918), Weimarer Republik (WR, 1918--1933), Nationalsozialismus (NS, 1933--1945). 
In the case of the Bundesrepublik (1949--2020), 
we divided the data into slices  of contiguous time periods in which one of the two major German parties (Volksparteien, i.e., liberal-conservative CDU or social-democratic SPD) was in charge. 
Table \ref{table:statistics} shows statistics of our data. For the Bundestagsprotokolle, slices range from between $\sim$30 million to $\sim$66 million tokens. For the Reichstagsprotokolle, the NS era is severely restricted in size, congruent with the observation that the parliamentary processes during the NS time were largely abolished.\footnote{\scriptsize\url{https://www.bundestag.de/parlament/geschichte/parlamentarismus/drittes_reich}}

\vspace{1em}
\noindent
\textbf{Party distribution.} The number of parties in our historical time periods with active speakers ranges from 2 (during the NS period, including SPD and NSDAP) to 10 (during KR2).\footnote{In the original data, party affiliation is attached to the speaker names.} Of course, the party distribution may decisively affect our results, i.e., distribution of biases found. In this context, it is worth noting that certain parties were banned during the time span of our data. For example, the right-wing party NSDAP was forbidden after 1945. The communist party KPD was forbidden in 1956. Our data thus contains communist parties only during the time period of WR and CDU1.  

\begin{table}[!t]
    \caption{Labels for time periods, range of time periods, and sizes of the corresponding data slices.}
    \centering
    \begin{tabular}{l|crr}
    \toprule
    \textbf{Period} & \textbf{Years} 
    & \textbf{Tokens} \\ \midrule
         KR1 &	1867-1890 &	
         40,585,912 
         \\
KR2	& 1890-1918 &	
         77,175,976\\ 
WR	&1918-1933 &	
         35,838,922 
         \\
NS	&1933-1942 &	
         230,018 \\ 
CDU1&	1949-1969 &	
     43,337,027 \\ 
SPD1&	1969-1982 &	
     35,208,879 \\ 
CDU2&	1982-1998 &	
     55,451,433 \\
SPD2&	1998-2005 &	
     28,614,189 \\ 
CDU3&	2005-2020 &	
     66,192,033 \\
         \bottomrule 
    \end{tabular}
    \label{table:statistics}
\end{table}

\begin{figure*}
    \centering
    \includegraphics[scale=0.25]{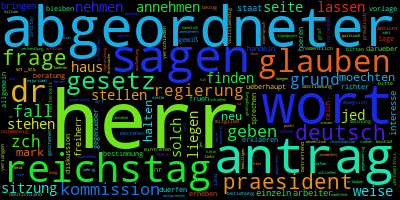}
    \includegraphics[scale=0.25]{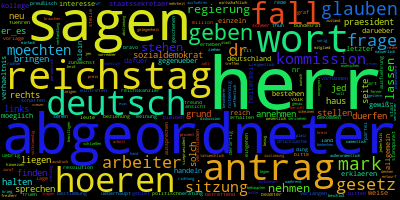}
    \includegraphics[scale=0.25]{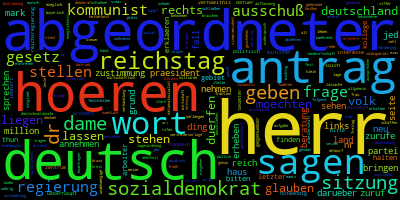}
    \includegraphics[scale=0.25]{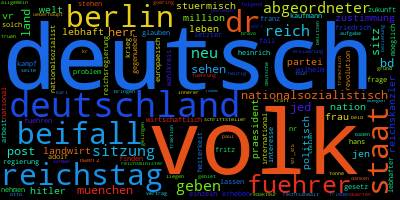}\\
    \includegraphics[scale=0.22]{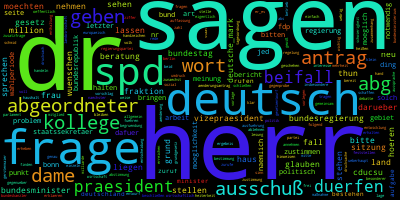}
    \includegraphics[scale=0.22]{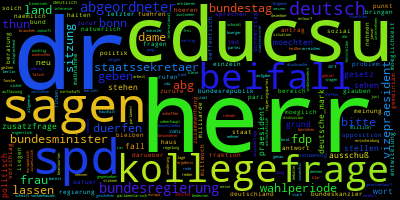}
    \includegraphics[scale=0.22]{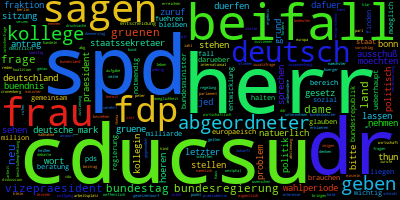}
    \includegraphics[scale=0.22]{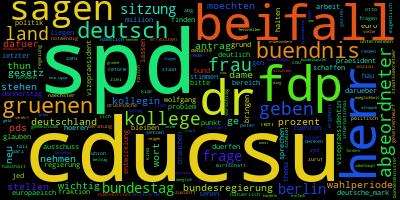}
    \includegraphics[scale=0.22]{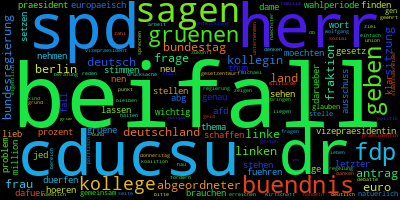}
    \caption{Word clouds showing relative frequency of words in text data for our nine periods: KR1, KR2, WR, NS (top, from left to right) and CDU1, SPD1, CDU2, SPD2, CDU3 (bottom).}
    \label{fig:wordclouds}
\end{figure*}

\vspace{1em}
\noindent
\textbf{Corpus analysis.} In Figure \ref{fig:wordclouds}, we show so-called `word clouds' for each slice in our timeline (1867--2020).\footnote{We use \url{https://github.com/amueller/word_cloud} for drawing.} The word clouds graphically illustrate the frequency of words in text data, assigning more frequent words larger font sizes (we removed stop words and only consider words with a minimum frequency of 50 occurrences). As expected, vocabulary relevant for parliamentary debates is prominent in our textual material, e.g., `Abgeordneter' (\emph{representative}), `Herr' (\emph{mr.}), `Antrag' (\emph{petition}). Names for political parties (`SPD', `FDP', `CDU/CSU') become more frequent in more recent time periods.
Interestingly, the word `deutsch' (\emph{German}) becomes more prominent from KR1 to KR2 to WR and then peaks in the NS period, where the word `Volk' (ideologically-loaded \emph{people} as nation) along with `Führer' (\emph{leader}, used as title for Nazi dictator Adolf Hitler) and `Berlin' also become very prominent. The trend for `deutsch' then reverses with the onset of the Bundesrepublik, i.e., after 1949, thus intuitively aligning with general understatement of national and cultural identity in Germany after World War II.

\begin{figure}
    \centering
    \includegraphics[scale=.45]{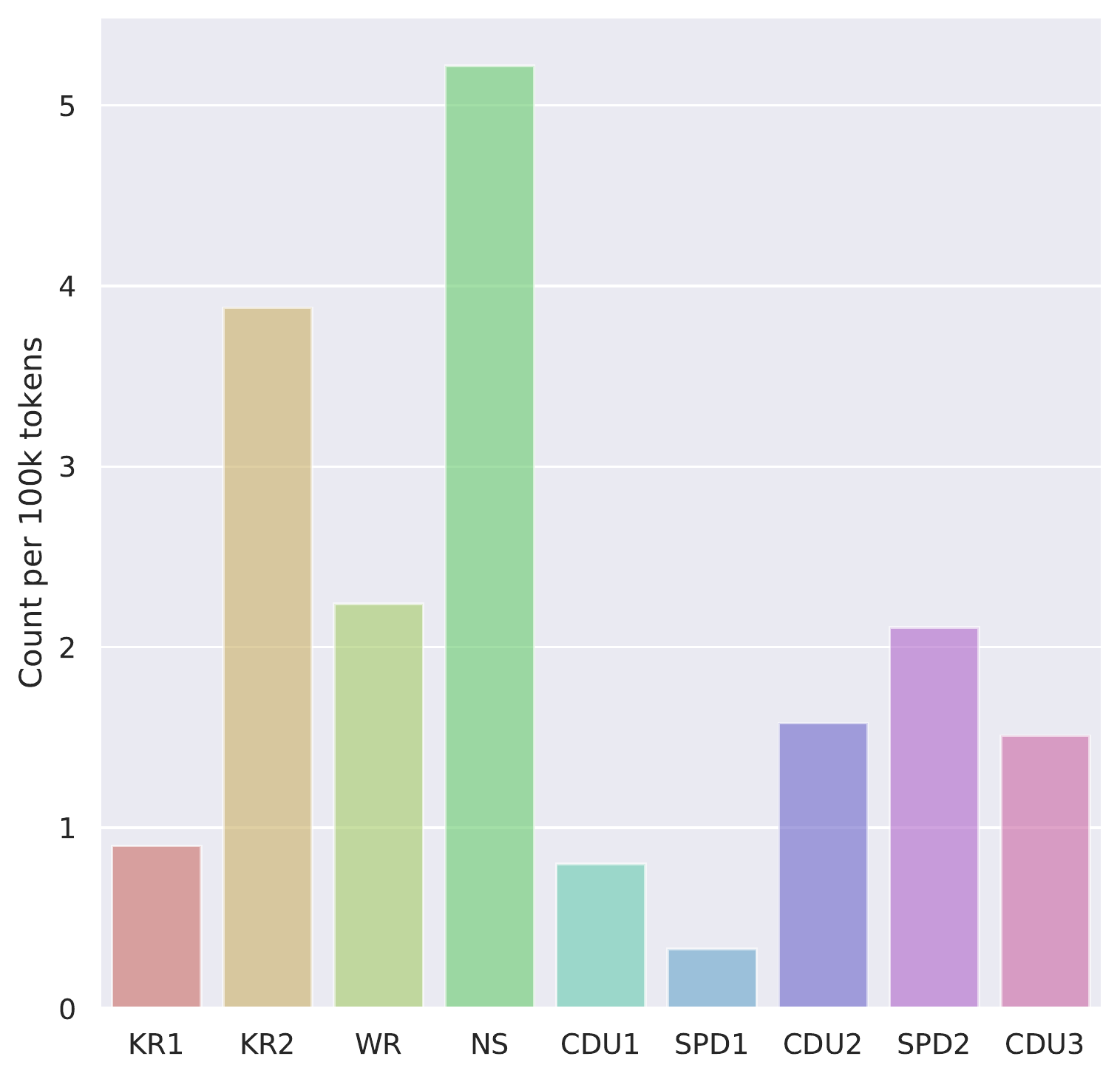}
    \caption{Histogram summarizing the number of occurrence of the German word `Jude' (\emph{jew}) per 100k tokens across different time slices of our corpus.}
    \label{fig:jude}
\end{figure}

\begin{figure}
    \centering
    \includegraphics[scale=0.95, trim=0.7cm 0cm 0.25cm 0cm, clip]{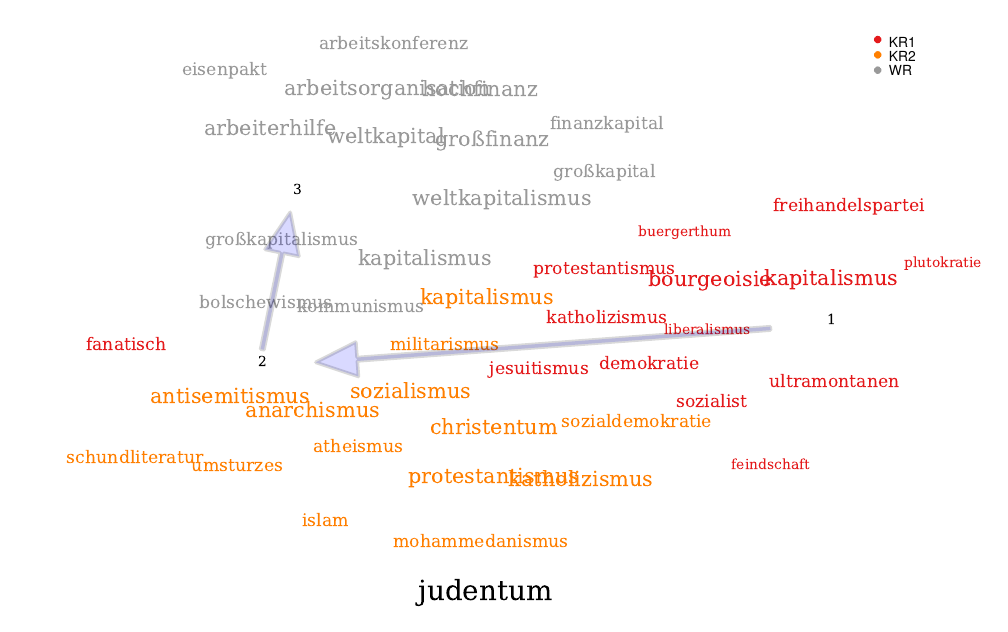}
    \caption{
    The semantic representations of the word `Judentum' (denoted by the numbers in chronological order, i.e., 1:KR1, 2:KR2, 3: WR) and their 15 nearest neighbors in the semantic space of each Reichstag period (red: KR1, orange:KR2, gray:WR).
    }
    \label{fig:semantic_change}
\end{figure}

A limitation of word clouds is that they can highlight only a few frequent words from the corpus. To also analyze trends for lower frequency and highly relevant words, e.g., `Jude' (\emph{jew}) -- due to its central relevance for one of the two kinds of bias that we focus on (i.e., antisemitism) -- we show a temporal curve in Figure \ref{fig:jude}. Like `deutsch', `Jude' peaks in the NS period, with a frequency of almost 6 per 100k tokens, then sharply declines in the post war period to regain prominence again after 1982, though not on the same level as before 1949 (disregarding KR1).  Finally, Figure \ref{fig:semantic_change} shows semantic shift \cite{eger-mehler-2016-linearity} of the word `Judentum' (\emph{Judaism}) in our data. The plot is obtained by semantically embedding  words from different time periods in a shared vector space using the model of \cite{Di_Carlo_Bianchi_Palmonari_2019} and then tracking in the semantic space how the word changes its fine-granular meaning over time periods using t-SNE \cite{vanDerMaaten2008} to visualize the embeddings (cf.\ also HistWords \cite{hamilton-etal-2016-diachronic}). In this case, for instance, we see how `Judentum' seemingly takes on (putatively negative) associations of `Bolschewismus' (\emph{Bolshevism}) and `Weltkapitalismus' (\emph{world capitalism}) in the time of the Weimar Republic. 

\section{Capturing Political Biases: Methodology}

\begin{figure}[t]
         \centering
         \includegraphics[width=1.0\linewidth,trim=1.5cm 3.5cm 4.5cm 1.5cm]{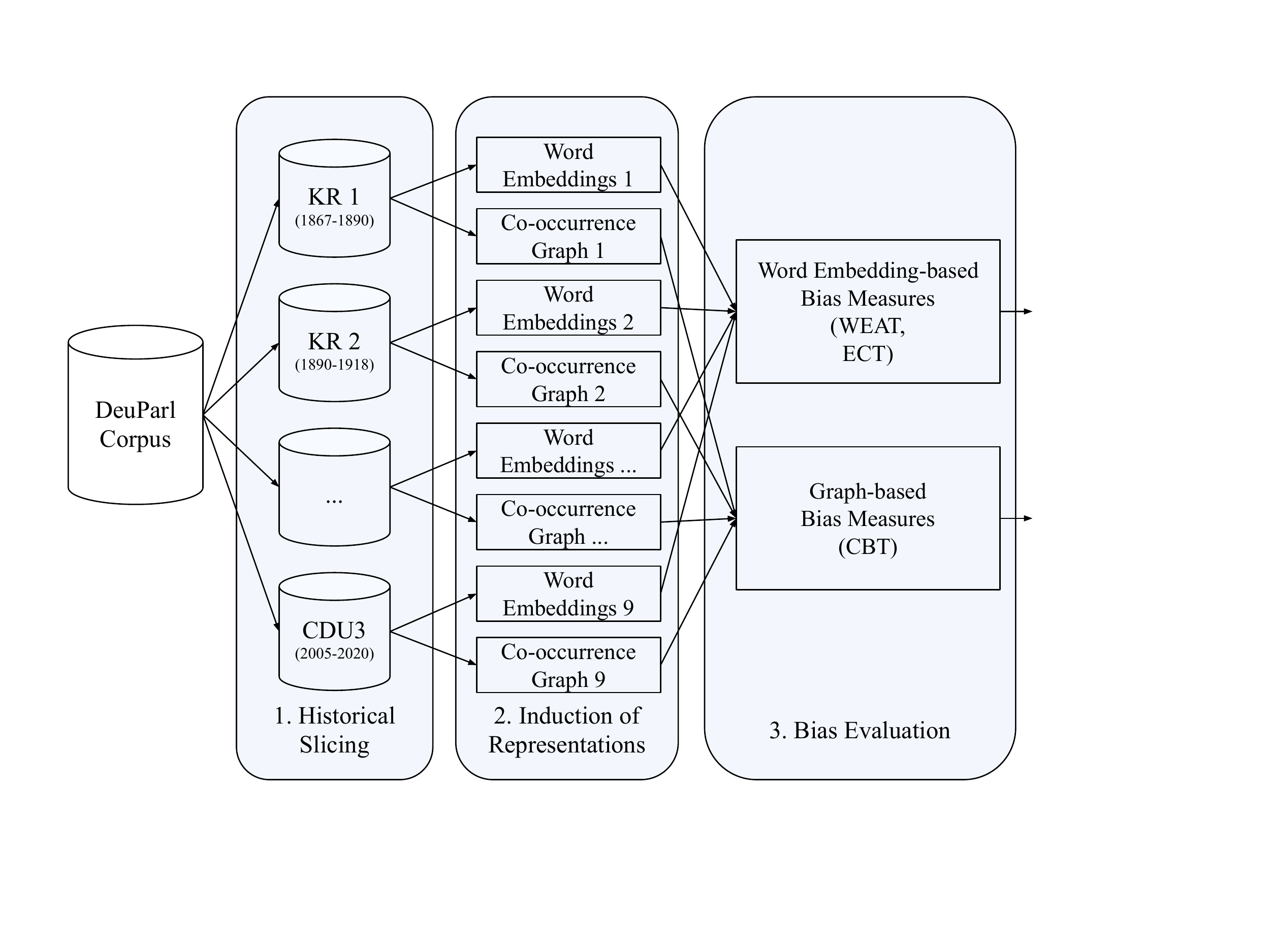}
         \caption{Overview of our bias evaluation pipeline.}
         \label{fig:pipeline}
\end{figure}
Our bias evaluation pipeline is illustrated in Figure~\ref{fig:pipeline}. The methodology consists of three main steps. We first slice \corpus according to the dimension of interest: in this paper, our focus is on politically well-defined historical periods. We then induce computational text representations, namely (1) dense distributional word vector spaces (i.e., word embeddings) and (2) word co-occurrence graphs, for each of the corpus slices. Finally, given human-designed bias specifications for \textit{antisemitism} and \textit{anti-communism}, we compute bias scores with several measures on the basis of either word embedding spaces or word coocurrence graphs induced in the previous step. We next describe our bias specifications, followed by the detailed description of our word embedding-based and co-occurrence graph-based approaches for quantifying the bias effects across different time periods.

\subsection{Bias Specifications}
In order to quantify the degree of presence of some societal bias (e.g., racial, religious, political, or gender bias), one must first specify (i.e., formalize) that bias. In this work, we follow the established body of work on measuring biases in word embedding spaces \cite{caliskan2017semantics,dev2019attenuating,lauscher2020general} and adopt the so-called \textit{explicit bias specification} $B_E=(T_1, T_2, A_1, A_2)$, consisting of two target term sets ($T_1$ and $T_2$), between which a significantly different association with respect to two sets of attribute terms ($A_1$ and $A_2$) is expected to exist.  In this work, we investigate two dimensions of political biases, which arguably played the most important role in (most recent) German history: (1) antisemitic bias, i.e., the bias between the dominant Christian group ($T_1$) and the targeted Jewish group ($T_2$),\footnote{We decided to define Judaism along a religious dimension, rather than along a racial dimension, as often perceived e.g.\ during the NS times. Thus, we contrasted Jews to Christians rather than Germans or Aryans. We speculate that this could lead to an underestimation of antisemitic bias.}
and (2) anti-communist bias, as the bias between the mainstream political conservatism ($T_1$) and the targeted communism ($T_2$). We expect the diachronic analysis of the degree of presence (or lack thereof) of these two bias types to point to ideological shifts in German political history. In what follows, we describe concrete bias specifications for each of the two dimensions of analysis. Table \ref{tab:specifications} provides an overview on all bias specifications employed in our study.

\vspace{1em}
\noindent
\textbf{Antisemitism.} We quantify the presence of antisemitism as the difference in association between `Christian' terms ($T_1$) and `Jewish ' terms ($T_2$) with respect to some attribute term sets $A_1$ (e.g., terms of positive sentiment like \textit{love}) and $A_2$ (e.g., negative sentiment terms like \textit{hate}). We obtain the terms for $T_1$ (i.e., Christian terms) by translating the Christian bias terms introduced by \cite{Garg.2018} from English to German (e.g., \textit{church} $\rightarrow$ \textit{Kirche}). We directly translated each term from the original English set and then judged the suitability of the translation for our use-case of diachronic bias analyses, resulting in minor modifications with respect to the original set. For instance, we discard the term `Kreuz' (\textit{cross}) due to its participation in the collocation `Eisernes Kreuz', one of the highest Prussian military accolades. 
Besides, we added to $T_1$ the following terms: `Pfarrer' (\textit{pastor}), `Ostern' (\emph{Easter}) and `Bibel' (\emph{Bible}). 
We devised the target term set for Judaism from scratch, starting from central terms that clearly denote this religion, e.g., `j{\"u}disch' (\textit{Jewish}) and `Jude' (\textit{Jew}). 

We observe the differences between Christian and Jewish target sets through seven different \textit{views}, each of which is defined with one pair of attribute sets ($A_1$, $A_2$). Views b--g) correspond to five bias lines identified by \cite{Tripodi.}: 

\vspace{.5em}
\begin{itemize}[leftmargin=4mm]
    \item[a)] \textbf{Sentiment.} We rely on the widely-used attribute set of pleasant ($A_1$) and unpleasant ($A_2$) terms introduced by \cite{caliskan2017semantics} and translated to German by \cite{lauscher-glavas-2019-consistently}; 
   \item[b)] \textbf{Religious.} Terms capturing a religious view of what is ``good'' and ``bad'', e.g., `Gläubige' (\textit{believers}, $A_1$); 
   \item[c)] \textbf{Economic.} Terms capturing the ``jew as greedy'' stereotype; 
   \item[d)] \textbf{Patriotic.} Nationally-loaded terms, e.g., `patriotisch' (\emph{patriotic}, $A_1$), `fremd' (\textit{foreign}, $A_2$); 
   \item[e)] \textbf{Racial.} Racist-related terminology capturing the idea of superior and inferior races;
   \item[f)] \textbf{Conspirational.} Terms related to ``jew world conspiracy' theories, e.g., `loyal' ($A_1$), betr\"{u}gerisch' (\textit{deceitful}, $A_2$); 
   \item[g)] \textbf{Ethic}. Includes ethically-loaded terms describing what is ``good'' and ``bad'', e.g., `tugendhaft' (\textit{virtuous}, $A_1$). 
\end{itemize}
\vspace{.5em}

For each of these views, terms in $A_2$ have been recognized by \cite{Tripodi.} to be commonly used in constructions expressing negative stereotypes about Jews: for example, the negatively biased view of greedy and economically influential Jews symbolized by the Rothschilds (economic); the Jews involved in political machinations and secret plots symbolized by George Soros (conspiratorial); or the Nazi myth of Jews as distinct and inferior race (racial).

\vspace{1em}
\noindent
\textbf{Anti-communism.} In order to capture anti-communist bias, we devise two target term sets representing mainstream political conservatism ($T_1$) and the targeted communism ($T_2$). To this end, we resort to political and historical literature \cite{schildt1998konservatismus}. For both bias specifications, we account for linguistic changes across the analyzed range of $153$ years by adapting the bias specifications after $1949$ (starting with the German Bundestag). Here, we focus on three views:

\vspace{.5em}
\begin{itemize}[leftmargin=4mm]
    \item[a)] \textbf{Sentiment.} We employ the same set of terms capturing general sentiment as above;
    \item[b)] \textbf{Politics.} Terms relating to a political view of ``good'' and ``bad'', e.g., `wirksam' (\textit{effective}, $A_1$)
    \item[c)] \textbf{Propaganda.} Terms capturing anti-communist propaganda.
\end{itemize}

\begin{table*}[]
    \caption{Bias specifications for representing antisemitism and anti-communism employed in our study. Superscript $^-$ indicates terms which we remove and superscript $^+$ indicates terms which we add for tests on historical slices starting from $1949$ (German Bundestag).}
    \centering
    {\small
    \begin{tabularx}{1.0\textwidth}{l X}
         \toprule
         Bias Type/View & Specification terms \\
         \midrule
         \multirow{4}{*}{\textbf{Antisemitism}} & \textbf{$T_1$\,(Christentum)}: \textit{Taufe}, \textit{Katholizismus}, \textit{Christentum}, \textit{evangelisch}, \textit{Evangelium}, \textit{Jesus}, \textit{Christ}, \textit{christlich}, \textit{katholisch}, \textit{Kirche}, \textit{Pfarrer}, \textit{Ostern}, \textit{Bibel}   \\
         
         & \textbf{$T_2$\,(Judentum)}: \textit{Rabbiner}, \textit{Synagoge}, \textit{koscher}, \textit{Talmud}, \textit{orthodox}, \textit{Judentum}, \textit{Jude}, \textit{jüdisch}, \textit{Mose}, \textit{mosaisch}, \textit{Israel}, \textit{Abraham}, \textit{zionistisch}, \textit{israelitisch}, \textit{Israelis} \\ \cline{2-2} 
         
         \multirow{4}{*}{View:\,\textit{Sentiment}} & \textbf{$A_1$\,(+)}: \textit{streicheln}, \textit{Freiheit}, \textit{Gesundheit}, \textit{Liebe}, \textit{Frieden}, \textit{Freude}, \textit{Freund}, \textit{Himmel}, \textit{loyal}, \textit{Vergnügen}, \textit{Diamant}, \textit{sanft}, \textit{ehrlich}, \textit{glücklich}, \textit{Regenbogen}, \textit{Geschenk}, \textit{Ehre}, \textit{Wunder}, \textit{Sonnenaufgang}, \textit{Familie}, \textit{Lachen}, \textit{Paradies}, \textit{Ferien} \\
         
         & \textbf{$A_2$\,(-)}: \textit{Missbrauch}, \textit{Absturz}, \textit{Schmutz}, \textit{Mord}, \textit{Krankheit}, \textit{Tod}, \textit{Trauer}, \textit{vergiften}, \textit{stinken}, \textit{Angriff}, \textit{Katastrophe}, \textit{Haß}, \textit{verschmutzen}, \textit{Tragödie}, \textit{Scheidung}, \textit{Gefängnis}, \textit{Armut}, \textit{häßlich}, \textit{Krebs}, \textit{töten}, \textit{faul}, \textit{erbrechen}, \textit{Qual} \\ \cdashline{2-2} 
         
         \multirow{4}{*}{View:\,\textit{Patriotism}} & \textbf{$A_1$ (+)}: \textit{patriotisch}, \textit{vaterlandsliebe}, \textit{volksbewußtsein}, \textit{volksgeist}, \textit{germanische}, \textit{deutschnational}, \textit{nationalbewußstsein}, \textit{vaterländisch}, \textit{reichstreu}, \textit{nationalgesinnt}, \textit{nationalstolz}, \textit{königstreu}, \textit{staatstreu}, \textit{nationalgefühl}, \textit{Volkstum$^+$}, \textit{Patriotismus$^+$}, \textit{Patriot$^+$} \\
         
         & \textbf{$A_2$\,(-)}: \textit{nichtdeutsch}, \textit{fremdländisch}, \textit{fremd}, \textit{undeutsch}, \textit{vaterlandslos}, \textit{reichsfeind}, \textit{landesverräter}	\textit{reichsfeindlich}, \textit{unpatriotisch}, \textit{antideutsch}, \textit{deutschfeindlich}, \textit{umstürzler}, \textit{staatsfeindlich}, \textit{dissident}, \textit{heimatlos}, \textit{separatistisch$^+$}, \textit{staatenlos$^+$}  \\ \cdashline{2-2} 
         
         \multirow{2}{*}{View:\,\textit{Economy}} & \textbf{$A_1$ (+)}: \textit{geben}, \textit{Großzügigkeit}, \textit{großzügig}, \textit{selbstlos}, \textit{genügsam}, \textit{Großmut}, \textit{uneigennützig}, \textit{sparsam}, \textit{Proletariat}, \textit{Armut}, \textit{Industriearbeiter} \\
         
         & \textbf{$A_2$\,(-)}: \textit{nehmen}, \textit{Gier}, \textit{gierig}, \textit{egoistisch}, \textit{habgierig}, \textit{Habsucht}, \textit{eigennützig}, \textit{verschwenderisch}, \textit{Bourgeoisie}, \textit{Wohlstand}, \textit{Bankier}, \textit{Wucher$^+$} \\ \cdashline{2-2} 
         
         \multirow{4}{*}{View:\,\textit{Conspiracy}} & \textbf{$A_1$ (+)}: \textit{loyal}, \textit{Kamerad}, \textit{Ehrlichkeit}, \textit{ersichtlich}, \textit{aufrichtig}, \textit{vertrauenswürdig}, \textit{wahr}, \textit{ehrlich}, \textit{unschuldig}, \textit{freundschaftlich}, \textit{hell}, \textit{zugänglich}, \textit{machtlos}, \textit{ohnmacht}, \textit{untertan}  \\
         
         & \textbf{$A_2$\,(-)}: \textit{illoyal}, \textit{Spitzel}, \textit{Verrat}, \textit{geheim}, \textit{hinterlistig}, \textit{betrügerisch}, \textit{unwahr}, \textit{zweifelhaft}, \textit{Verbrecher}, \textit{bedrohlich}, \textit{dunkel}, \textit{geheimnis}, \textit{einflussreich}, \textit{weltmacht}, \textit{herrschaft}, \textit{verschwoerung} \\ \cdashline{2-2} 
         
         \multirow{2}{*}{View:\,\textit{Religion}} & \textbf{$A_1$ (+)}: \textit{Gläubige}, \textit{geistlich}, \textit{Engel}, \textit{heilig}, \textit{fromm}, \textit{geheiligt}, \textit{göttlich}, \textit{ehrwürdig}, \textit{treu}, \textit{gläubig}, \textit{religiös} \\
         
         & \textbf{$A_2$\,(-)}: \textit{Atheist}, \textit{weltlich}, \textit{Teufel}, \textit{irdisch}, \textit{atheistisch}, \textit{heidnisch}, \textit{gottlos}, \textit{verflucht}, \textit{untreu}, \textit{ungläubig}, \textit{irreligiös}, \textit{Gotteslästerung} \\ \cdashline{2-2} 
         
         \multirow{4}{*}{View:\,\textit{Race}} & \textbf{$A_1$ (+)}: \textit{normal}, \textit{überlegenheit}, \textit{gleichheit}, \textit{angenehm}, \textit{freundlich}, \textit{ehrenwert}, \textit{sympathie}, \textit{akzeptiert}, \textit{besser}, \textit{national}, \textit{rein}, \textit{überlegen}, \textit{sauber}, \textit{ehrenhaft} \\
         
         & \textbf{$A_2$\,(-)}: \textit{seltsam}, \textit{unterlegenheit}, \textit{ungleichheit}, \textit{unangenehm}, \textit{boshaft}, \textit{schändlich}, \textit{hass}, \textit{abgelehnt}, \textit{schlechter}, \textit{fremdländisch}, \textit{unrein}, \textit{unterlegen}, \textit{schmutzig}, \textit{verseucht}, \textit{schädlich}, \textit{niederträchtig} \\ \cdashline{2-2} 
         
         \multirow{3}{*}{View:\,\textit{Ethics}} & \textbf{$A_1$ (+)}: \textit{bescheiden}, \textit{sittlich}, \textit{anständig}, \textit{tugendhaft}, \textit{charakterfest}, \textit{würdig}, \textit{treu}, \textit{moralisch}, \textit{ehrlich}, \textit{gesittet}, \textit{gewissenhaft}, \textit{vorbildlich} \\
         
         & \textbf{$A_2$\,(-)}: \textit{unbescheiden}, \textit{unsittlich}, \textit{unanständig}, \textit{lüstern}, \textit{korrupt}, \textit{unwürdig}, \textit{untreu}, \textit{unmoralisch}, \textit{unehrlich}, \textit{verdorben}, \textit{gewissenlos}, \textit{barbarisch} \\ \midrule

         \multirow{2}{*}{\textbf{Anti-communism}}   
         & \textbf{$T_1$\,(Conservatism)}: \textit{Konservatismus}, \textit{Tradition}, \textit{Geschichte}, \textit{Christentum$^{-}$}, \textit{Adel$^{-}$}, \textit{Monarchie$^{-}$}, \textit{Mittelalter$^{-}$}, \textit{Stände$^{-}$}, \textit{Werte}, \textit{Moral}, \textit{König$^{-}$}, \textit{Kaiser$^{-}$}, \textit{Hierarchie}, \textit{Identität}, \textit{Kontinuität}, \textit{Sicherheit}, \textit{Grundbesitz$^{-}$}, \textit{Autorität}, \textit{Legitimität},
\textit{Ordnung}, \textit{Religion$^{-}$}, \textit{Kirche$^{-}$}, \textit{Erhaltung}, \textit{Treue$^{-}$}, \textit{Tugend$^{-}$}, \textit{Bräuche}, \textit{Sitten}, \textit{Bewahrung}, \textit{Gottesgnadentum$^{-}$},
\textit{Ständeordnung$^{-}$}, \textit{Restauration$^{-}$}, \textit{Bürger$^{+}$}, \textit{Bürgertum$^{+}$},  \textit{Regierung$^{+}$}, \textit{Wertordnung$^{+}$}, \textit{Bürgerlichkeit$^{+}$}, \textit{Stabilität$^{+}$}, \textit{Wohlstand$^{+}$} \\
         
& \textbf{$T_2$\,(Communism)}: \textit{Sozialismus$^{-}$}, \textit{Kommunismus}, \textit{Proletariat}, \textit{Arbeiter$^{-}$}, \textit{Klassengesellschaft}, \textit{Klasse$^{-}$}, \textit{Revolution}, \textit{Aufklärung$^{-}$}, \textit{Gemeinschaft$^{-}$},
\textit{Gerechtigkeit$^{-}$}, \textit{Armut$^{-}$}, \textit{Kapital}, \textit{Gleichheit$^{-}$}, \textit{Chancen$^{-}$}, \textit{Freiheit$^{-}$}, \textit{Arbeiterklasse$^{-}$}, \textit{Solidarität$^{-}$}, \textit{Partei$^{-}$}, \textit{Verstaatlichung},
\textit{Gewerkschaft$^{-}$}, \textit{Marx}, \textit{Engels}, \textit{Vergesellschaftung}, \textit{Gemeineigentum}, \textit{Widerstand}, \textit{Kollektivierung},\textit{Arbeiterbewegung$^{-}$},
\textit{Aufstand$^{-}$}, \textit{Lenin$^{+}$}, \textit{Planwirtschaft$^{+}$}, \textit{Klassenkampf$^{+}$}, \textit{Proletariat$^{+}$}, \textit{Revolution$^{+}$}, \textit{Produktionsmittel$^{+}$}, \textit{Diktatur$^{+}$}, \textit{Bolschewiki$^{+}$} ,\textit{Oktoberrevolution$^{+}$},\textit{Räte$^{+}$}, \textit{Sowjetunion$^{+}$}\\ 

\cline{2-2} 

        \multirow{4}{*}{View:\,\textit{Sentiment}} & \textbf{$A_1$\,(+)}: \textit{streicheln}, \textit{Freiheit}, \textit{Gesundheit}, \textit{Liebe}, \textit{Frieden}, \textit{Freude}, \textit{Freund}, \textit{Himmel}, \textit{loyal}, \textit{Vergnügen}, \textit{Diamant}, \textit{sanft}, \textit{ehrlich}, \textit{glücklich}, \textit{Regenbogen}, \textit{Geschenk}, \textit{Ehre}, \textit{Wunder}, \textit{Sonnenaufgang}, \textit{Familie}, \textit{Lachen}, \textit{Paradies}, \textit{Ferien} \\
         
         & \textbf{$A_2$\,(-)}: \textit{Missbrauch}, \textit{Absturz}, \textit{Schmutz}, \textit{Mord}, \textit{Krankheit}, \textit{Tod}, \textit{Trauer}, \textit{vergiften}, \textit{stinken}, \textit{Angriff}, \textit{Katastrophe}, \textit{Haß}, \textit{verschmutzen}, \textit{Tragödie}, \textit{Scheidung}, \textit{Gefängnis}, \textit{Armut}, \textit{häßlich}, \textit{Krebs}, \textit{töten}, \textit{faul}, \textit{erbrechen}, \textit{Qual} \\ \cdashline{2-2} 
 
         \multirow{4}{*}{View:\,\textit{Politics}} & \textbf{$A_1$\,(+)}:  \textit{sozial}, \textit{progressiv}, \textit{gemeinschaftlich}, \textit{gemeinsam}, \textit{zivilisiert}, \textit{bewährt}, \textit{wirksam}, \textit{etabliert}, \textit{demokratisch},
\textit{hoch}, \textit{möglich}, \textit{fortschrittlich}, \textit{gemäßigt}, \textit{machbar}, \textit{realistisch}, \textit{früh}, \textit{kontinuierlich}, \textit{legitim}, \textit{verlässlich},
\textit{aufrichtig}, \textit{intellektuell}, \textit{sicher}, \textit{Sicherheit}, \textit{Fortschritt}, \textit{pragmatisch}, \textit{Vertrauen}, \textit{Wandel$^+$}, \textit{sachlich$^+$}, \textit{Gewinn$^+$}, \textit{fähig$^+$}\\
         
         & \textbf{$A_2$\,(-)}: \textit{unsozial}, \textit{radikal}, \textit{extrem}, \textit{gefährlich}, \textit{gefährdend}, \textit{niedrig}, \textit{nieder}, \textit{unmöglich}, \textit{undemokratisch}, \textit{unrealistisch},
\textit{spät}, \textit{unlegitim}, \textit{Gefahr}, \textit{unehrlich}, \textit{unaufrichtig}, \textit{unintellektuell}, \textit{unsicher}, \textit{schwer}, \textit{schwierig}, \textit{Misstrauen},  \textit{Stillstand$^+$},  \textit{Skandal$^+$}, \textit{skandalös$^+$}, \textit{Zukunft$^+$}, \textit{unsachlich$^+$}, \textit{Verlust$^+$}, \textit{unfähig$^+$}  \\ \cdashline{2-2} 
 
\multirow{4}{*}{View:\,\textit{Propaganda}} & \textbf{$A_1$\,(+)}:  \textit{Kamerad$^-$}, \textit{Kameraden$^-$}, \textit{Kameradschaft$^-$}, \textit{kameradschaftlich$^-$}, \textit{Vaterland$^-$}, \textit{Patriot$^-$}, \textit{Ehre}, \textit{ehrlich}, \textit{Einsatz}, \textit{Untertan$^-$}, \textit{rein$^-$}, \textit{wir}, \textit{Heimat$^-$}, \textit{deutsch}, \textit{Deutschland$^-$}, \textit{Truppe$^-$}, \textit{Nationalstolz$^-$}, \textit{patriotisch}, \textit{Volk}, \textit{Befreiung$^-$}, \textit{Front$^-$}, \textit{Wahrheit}, \textit{wahr}, \textit{aufrichtig$^+$}, \textit{gemeinschaftlich$^+$}, \textit{Wertegemeinschaft$^+$}, \textit{Mitte$^+$}, \textit{Frieden$^+$}, \textit{Partnerschaft$^+$}, \textit{Integration$^+$}, \textit{Wandel$^+$} \\
         
& \textbf{$A_2$\,(-)}: \textit{Sabotage$^-$}, \textit{Saboteur$^-$}, \textit{Betrüger}, \textit{Betrug}, \textit{Gauner$^-$}, \textit{Schwindel$^-$}, \textit{Schwindler$^-$}, \textit{Parasit$^-$}, \textit{Volksfeind$^-$}, \textit{Reichsfeind$^-$}, \textit{undeutsch}, \textit{unpatriotisch}, \textit{reichsfeindlich}, \textit{Volksverräter$^-$}, \textit{Spion$^-$}, \textit{Bolschewist$^-$}, \textit{fremd}, \textit{unrein$^-$}, \textit{Kommunist}, \textit{Spitzel$^-$}, \textit{anders}, \textit{Lüge}, \textit{Lügner}, \textit{Dissident}, \textit{Feind}, \textit{Diktatur}, \textit{Verschwörung$^-$}, \textit{verschwörerisch$^-$}, \textit{unehrlich$^+$}, \textit{feindlich$^+$}, \textit{Schmarotzer$^+$}, \textit{Elite$^+$}, \textit{Kriminelle$^+$}, \textit{kriminell$^+$}\\ 
         \bottomrule
    \end{tabularx}}
    \label{tab:specifications}
\end{table*}

\subsection{Measuring Bias in Semantic Spaces.} Most existing measures for capturing bias in text corpora (e.g., \cite{bolukbasi2016man,caliskan2017semantics,Garg.2018,gonen2019lipstick,dev2019attenuating,lauscher-glavas-2019-consistently,lauscher2020general}, \emph{inter alia}), given an explicit bias specification, operate on word embeddings, namely dense vector representations of meaning in context \cite{mikolov2013distributed,pennington2014glove,bojanowski2017enriching}. In this work, we adopt two of the arguably most established bias measures based on word embeddings: Word Embedding Association Test (WEAT) \cite{caliskan2017semantics} and Embedding Coherence Test (ECT) \cite{dev2019attenuating}. Training reliable word embeddings, however, requires substantial amounts of text. Given that some of the temporal slices in our 
\corpus{} 
corpus are of fairly limited size (see Table \ref{table:statistics}), questions remain on whether the word embeddings induced from those slices can be sufficiently reliable. Because of this, we couple the embedding-based bias measures with a novel bias measure based on term co-occurrences and graph-based label propagation, which we dub Co-occurrence Bias Test (CBT).          

\vspace{1em}
\noindent
\textbf{Quantifying Bias in Word Embeddings.} Word embeddings are dense numeric vector representations of words that aim to capture word meaning insofar that words with similar meaning get assigned vectors that are close to each other in the vector space. While there exist several different well-known algorithms for inducing word embeddings from a given corpus (see, e.g., \cite{mikolov2013distributed,pennington2014glove,bojanowski2017enriching}), they all rely on a distributional hypothesis \cite{harris1954distributional} and, one way or the other, exploit the information about local word co-occurrences to derive dense semantic vectors of words, i.e., word embeddings.        

\vspace{1em}
\noindent
\emph{Word Embedding Association Test (WEAT).} Introduced by \cite{caliskan2017semantics}, the WEAT test is an adaptation of the Implicit Association Test (IAT) \cite{nosek2002harvesting,nosek2005understanding}. Whereas IAT measures biases based on response times of human subjects to provided stimuli, WEAT quantifies the biases using semantic similarities between word embeddings of the same stimuli (i.e., terms in our bias specifications, see Table \ref{tab:specifications}). Given some word embedding space and an explicit bias specification $B_E$=$(T_1, T_2, A_1, A_2)$, the WEAT test statistic is computed as the differential of the associativity between $T_1$ and $T_2$ w.r.t. $A_1$ and $A_2$, stemming from the mean similarity of their terms with terms in the attribute sets $A_1$ and $A_2$, respectively: 

{
\begin{equation*}
        s(B_E) = \sum_{t_1 \in T_1}{s(t_1, A_1, A_2)} - \sum_{t_2 \in T_2}{s(t_2, A_1, A_2)}\,
\end{equation*}}%

\noindent The association score $s$ for an individual target term $t\in T_i$ is obtained as: 
{
\begin{equation*}
    s(t, A_1, A_2) = \frac{1}{|A_1|}\sum_{a_1 \in A_1}{\hspace{-0.3em}\cos(\mathbf{t}, \mathbf{a_1})}  -  \frac{1}{|A_2|}\sum_{a_2 \in A_2}{\hspace{-0.3em}\cos(\mathbf{t}, \mathbf{a_2})} 
\end{equation*}}%

\noindent where $\mathbf{t}$, $\mathbf{a_1}$, $\mathbf{a_2}$ are word embeddings of terms $t$, $a_1$, and $a_2$, respectively and $\cos{(\mathbf{x}, \mathbf{y})} \in [-1,1]$ is the cosine of the angle enclosed by the vectors $\mathbf{x}$ and $\mathbf{y}$. %
We note that $s(B_E)$ is large, for example, when terms in $T_1$ are positively associated with attributes in $A_1$ and negatively with attributes in $A_2$ \emph{and} vice versa for $T_2$. In this case, there is a bias for $T_1$ to be associated with $A_1$ and $T_2$ to be associated with $A_2$.

The significance of the above statistic is computed by comparing $s(B_E)$ with the scores $s(B^*_E)$ obtained for permutations of $B_E$, $B^*_E = (T^*_1, T^*_2, A_1, A_2)$, where $T^*_1$ and $T^*_2$ are equally sized partitions of $T_1 \cup T_2$. The $p$-value of the test is the probability of $s(B^*_E) > s(B_E)$. Finally, the \textit{bias effect size}, i.e., the ``amount'' of bias, is computed as the normalized measure of separation between association distributions:
%
{
\begin{equation*}
\frac{\mu\hspace{-0.1em}\left(\{s(t_1, A_1, A_2)\}_{t_1 \in T_1}\right) - \mu\hspace{-0.1em}\left(\{s(t_2, A_1, A_2)\}_{t_2 \in T_2}\right)}{\sigma\left(\{s(t, A_1, A_2)\}_{t \in T_1 \cup T_2}\right)}
\end{equation*}}%

\noindent with $\mu$ as the mean value and $\sigma$ as the standard deviation.

\vspace{1em}
\noindent
\emph{Embedding Coherence Test (ECT).} The ECT test, originally proposed by \cite{dev2019attenuating} and later adjusted by \cite{lauscher2020general} quantifies the amount of explicit bias by comparing vectors of target sets $T_1$ and $T_2$, obtained by averaging the embeddings of their constituent terms, with the vectors from a single attribute set $A = A_1 \cup A_2$. 
We first compute the embeddings for target sets $T_1$ and $T_2$ by averaging the vectors of their terms:

\begin{equation*}
    \mathbf{\mu}_1 = \frac{1}{|T_1|}\sum_{t_1 \in T_1}{\mathbf{t}_1}; \hspace{0.5em} \hspace{0.5em} \mathbf{\mu}_2 = \frac{1}{|T_2|}\sum_{t_2 \in T_2}{\mathbf{t}_2}.
\end{equation*}

\noindent Next, for both $\mathbf{\mu}_1$ and $\mathbf{\mu}_2$ it computes the (cosine) similarities with vectors of all attribute terms $a \in A$: 

\begin{equation*}
    \mathbf{s}_1 = \left[\cos(\mathbf{\mu}_1, \mathbf{a}_i)\right]^{|A|}_{i = 1}; \hspace{0.5em} \mathbf{s}_2 = \left[\cos(\mathbf{\mu}_2, \mathbf{a}_i)\right]^{|A|}_{i = 1}.
\end{equation*}

\vspace{0.5em}

\noindent The two resulting vectors of similarity scores, $\mathbf{s}_1$ (for $T_1$) and $\mathbf{s}_2$ (for $T_2$) are used to obtain the final ECT score: the Spearman's correlation between the rank orders of $\mathbf{s}_1$ and $\mathbf{s}_2$. Unlike for WEAT where a larger bias effect implies larger bias, the higher the ECT correlation, the lower the bias.

\vspace{1em}
\noindent
\textbf{Measuring Bias with Co-occurrence Graphs.} Both ECT and WEAT crucially depend on the quality of the semantic information encoded in the underlying word embedding space, which typically require large text corpora. 
We now propose a novel bias measure based on word co-occurrence graphs and graph-based label propagation, specifically designed to be applicable to small corpora as well. 

We first compute scores between terms in the corpus 
indicating 
the level of their lexical association, based on their co-occurrence frequency. Concretely, we adopt Positive Pointwise Mutual Information (PPMI) as the measure of lexical association between terms. For a pair of words $(w_1, w_2)$, pointwise mutual information (PMI) is computed as the probability of joint occurrence $P(w_1, w_2)$ of $w_1$ and $w_2$ (in a coocurrence window of fixed size), normalized with the product of the probabilities of individual terms, $P(w_1)$ and $P(w_2)$:
\begin{align*}
    \text{PMI} = \log\frac{P(w_1, w_2)}{P(w_1)\cdot P(w_2)}
\end{align*}

\noindent A PMI score of $0$ (i.e., probability ratio of $1$) 
means 
two words appear together exactly as frequently as one would expect from their individual frequencies. Scores lower than zero imply a reliable lack of association between terms. This is why PPMI replaces negative PMI scores with $0$, i.e., $\text{PPMI} = \max\{0,\text{PMI}\}$. 

\vspace{1em}
\noindent
\emph{Co-occurrence Bias Test (CBT).} The PPMI scores computed between all pairs of terms 
effectively define an undirected weighted graph which can then be used for semi-supervised label propagation of scores -- from a small subset of nodes for which such scores are known to all other (unlabeled) nodes in the graph. We assign the labels to the nodes corresponding to the terms of the two attribute lists: the terms from the positive attribute list $A_1$ are assigned the score $1$, whereas the terms from the negative attribute lists are labeled with $0$. We next employ the semi-supervised graph-based label propagation algorithm named Harmonic Function Label Propagation (HFLP) \cite{zhu2003semi,zhu2009introduction,glavavs2017unsupervised} to induce the scores for all other nodes (i.e., terms) in the PPMI graph.  
Let $\mathbf{W}$ be the weighted adjacency matrix of the PPMI graph with $w_{ij}$ as the PPMI score between the $i$-th and $j$-th vocabulary term, and let $\mathbf{D}$ 
be the corresponding diagonal degree matrix with entries $\mathbf{D}_{ii}=\sum_j w_{ij}$ and $\mathbf{D}_{ij}=0$ for $i\neq j$. Assuming we index all labeled nodes (i.e., nodes corresponding to terms from $A_1$ and $A_2$) in $W$ before all unlabeled nodes, the Laplacian of the graph, i.e., $\mathbf{\Delta} = \mathbf{W} - \mathbf{D}$, can then be partitioned as:

\begin{equation*}
\mathbf{\Delta} = \begin{pmatrix}
\mathbf{\Delta_{ll}} & \mathbf{\Delta_{lu}} \\
\mathbf{\Delta_{ul}} & \mathbf{\Delta_{uu}}
\end{pmatrix}
\end{equation*}

\noindent Let $\mathbf{f}_l$ be the binary vector of labels of labeled nodes (i.e., terms from $A_1$ and $A_2$). HFLP then offers a closed form solution for the scores of unlabeled nodes: 

\begin{equation*}
    \mathbf{f}_u = -\mathbf{\Delta}_{uu}^{-1}\mathbf{\Delta}_{ul}\mathbf{f}_l.
\end{equation*}

\noindent The scores $f \in \mathbf{f}_u$ assigned to unlabeled nodes will then be in between $0$ and $1$. We are now interested in the scores assigned to the terms from the two target lists $T_1$ and $T_2$, respectively. Concretely, we are interested in whether the mean of the scores for terms in $T_1$ statistically significantly differs from the mean of the scores assigned to terms in $T_2$: if so, then we are observing a significant bias effect with respect to the attribute sets $A_1$ and $A_2$. To this end, we apply the Student's (unpaired, two-tailed) $t$-test and interpret the value of the $t$-statistic directly as the size bias effect: 

\begin{equation*}
    t(T_1, T_2) = \frac{\mathbf{\mu}_{T_1} - \mathbf{\mu}_{T_2}}{s\cdot \sqrt{\frac{1}{|T_1|} + \frac{1}{|T_2|}}} 
\end{equation*}

\noindent with $\mathbf{\mu}(T)$ as the mean score that HFLP assigned to the terms in $T$ and $s$ as the estimator of the pooled standard deviation of the two sets of target scores: 

\begin{equation*}
    s = \sqrt{\frac{(|T_1| - 1)\cdot s^2_{T_1} + (|T_2| - 1)\cdot s^2_{T_2}}{|T_1| + |T_2| - 2}}
\end{equation*}

\noindent where $s^2_{T_1}$ and $s^2_{T_2}$ are the variances of the scores induced with HFLP for terms in $T_1$ and $T_2$, respectively.

\section{Experiments}
We present our experiments on measuring bias in our
\corpus{} corpus of German parliamentary proceedings. 

\subsection{Experimental Setup}
For the embedding-based approach, we train Word2Vec CBOW models~\cite{mikolov2013distributed} on each historical slice independently.
However, here, we exclude the NS slice due to its small size. We employ the \emph{gensim} framework for inducing the embeddings with an embedding size of $200$ and the window size as well as  the minimum count for unigrams set to $5$. For computing the PPMI matrices, we set the window size to $5$ and the minimum term count to $10$ with the exception of the NS slice for which we set the threshold to $2$. 

\subsection{Results}\label{sec:results}
\begin{figure*}[t]
     \centering
     \begin{subfigure}[t]{0.31\textwidth}
         \centering
         \includegraphics[width=1.0\linewidth,trim=0.0cm 0cm 1.5cm 0cm]{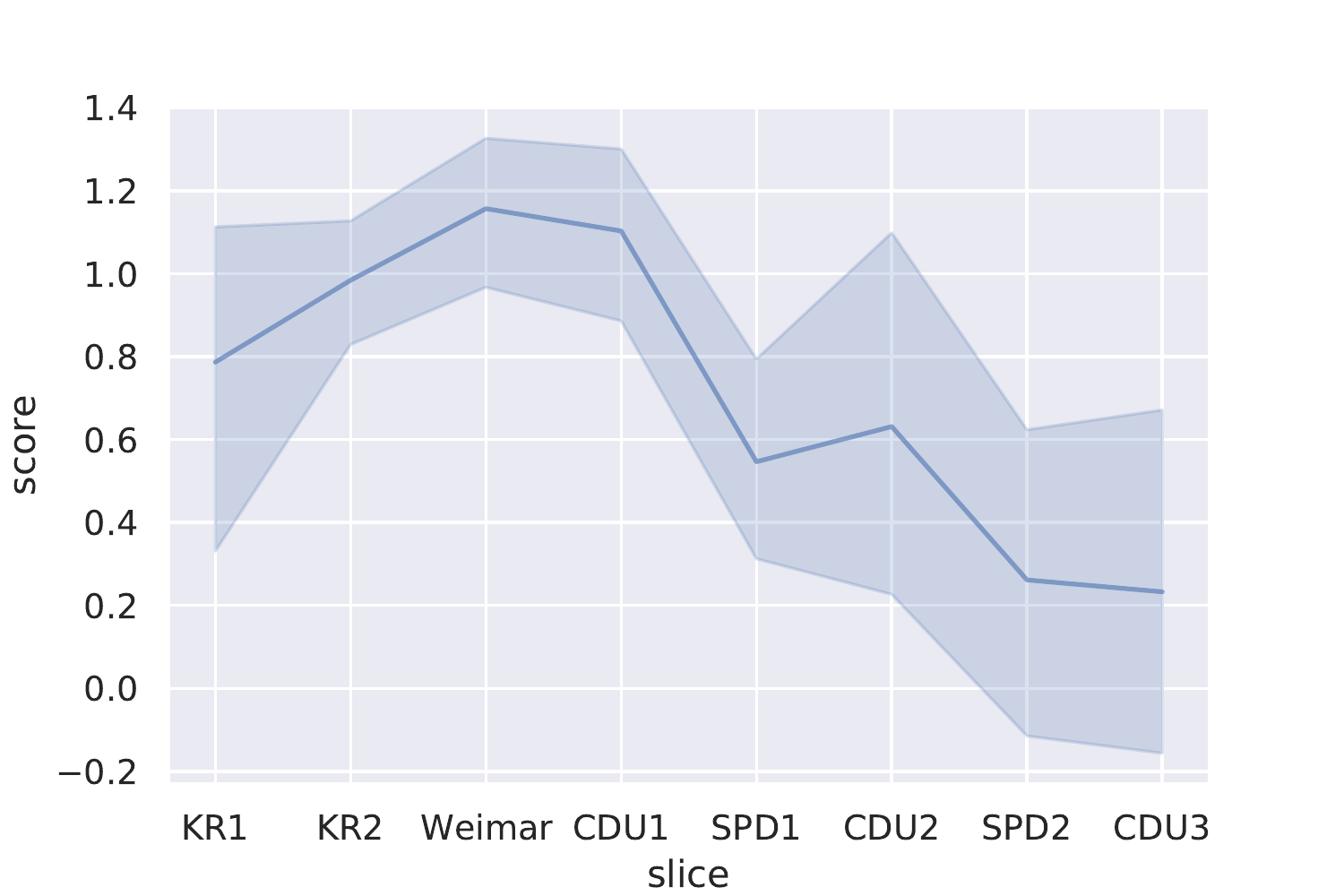}
         \caption{\footnotesize Aggregated WEAT scores (antisemitism).}
         \label{fig:aggr_weat_antisemitism}
    \end{subfigure}
    \begin{subfigure}[t]{0.31\textwidth}
         \centering
         \includegraphics[width=1.0\linewidth,trim=0.0cm 0cm 1.5cm 0cm]{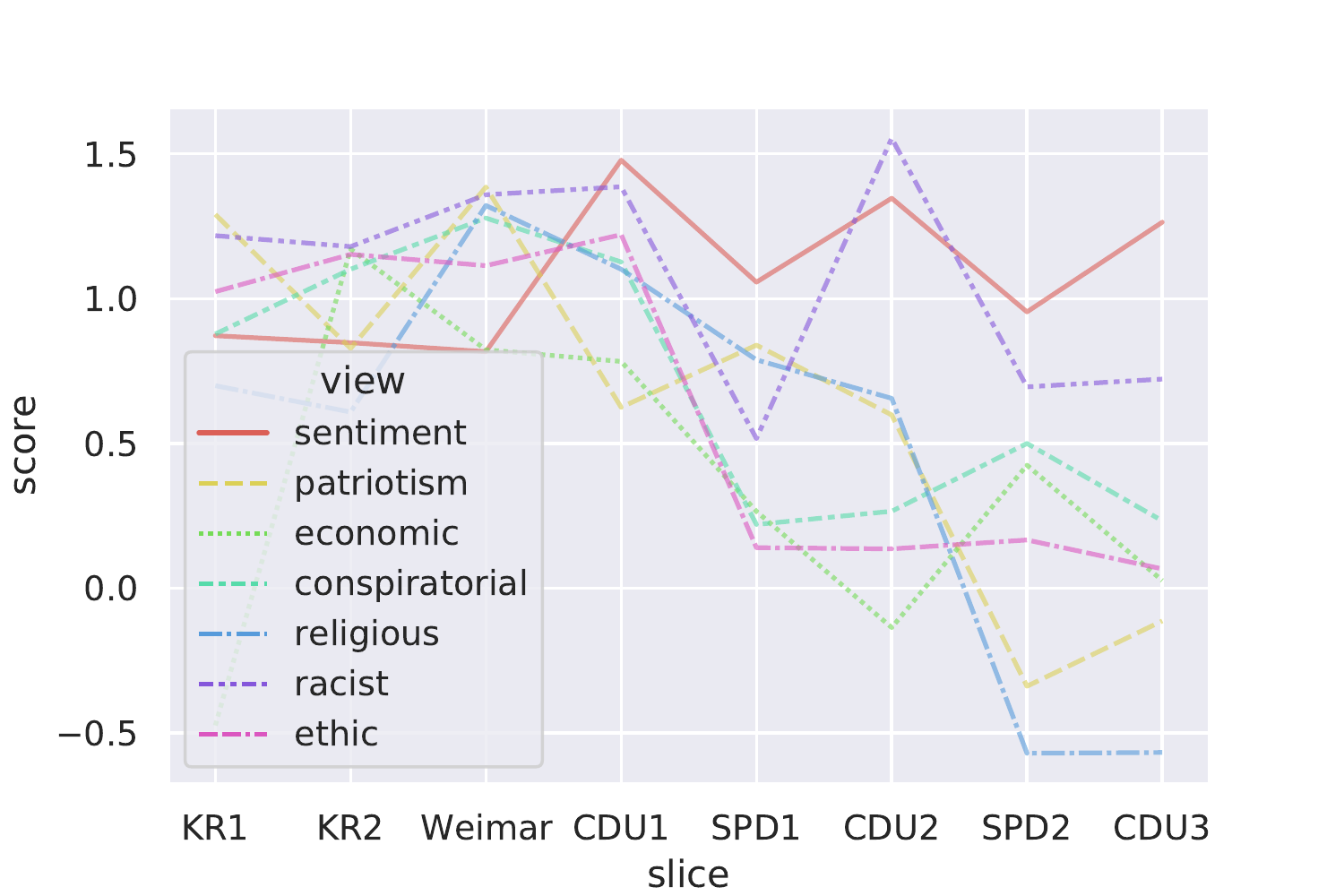}
         \caption{WEAT scores (antisemitism).}
         \label{fig:weat_antisemitism}
     \end{subfigure}
    \begin{subfigure}[t]{0.31\textwidth}
         \centering
         \includegraphics[width=1.01\linewidth,trim=0.0cm 0cm 1.5cm 0cm]{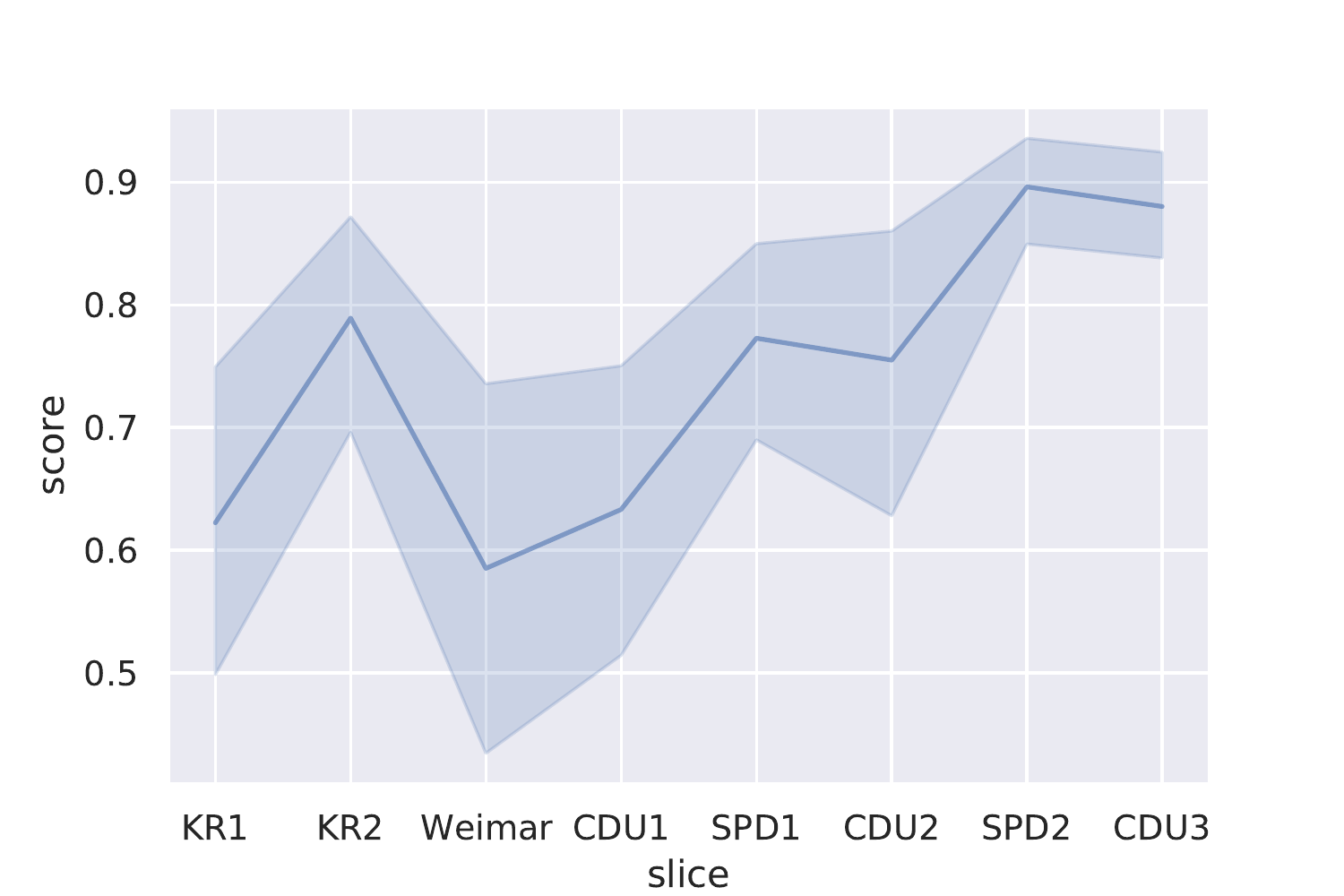}
         \caption{\footnotesize Aggregated ECT scores (antisemitism).}
         \label{fig:aggr_ect_antisemitism}
     \end{subfigure}
    \begin{subfigure}[t]{0.31\textwidth}
         \centering
         \includegraphics[width=1.0\linewidth,trim=0.0cm 0cm 1.5cm 0cm]{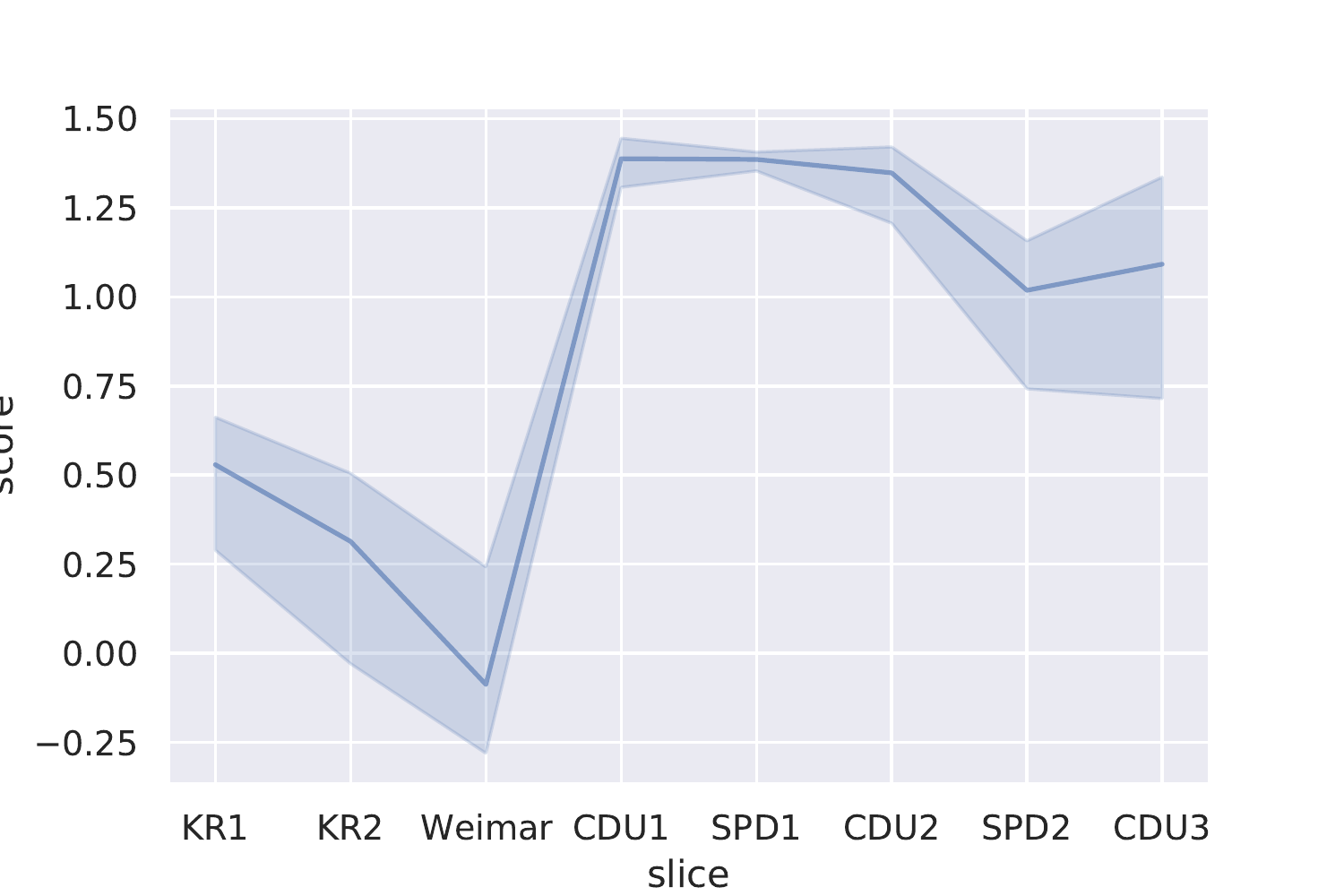}
         \caption{\footnotesize Aggregated WEAT scores (anti-communism).}
         \label{fig:aggr_weat_anticommunism}
     \end{subfigure}
    \begin{subfigure}[t]{0.31\textwidth}
         \centering
         \includegraphics[width=1.01\linewidth,trim=0.0cm 0cm 1.5cm 0cm]{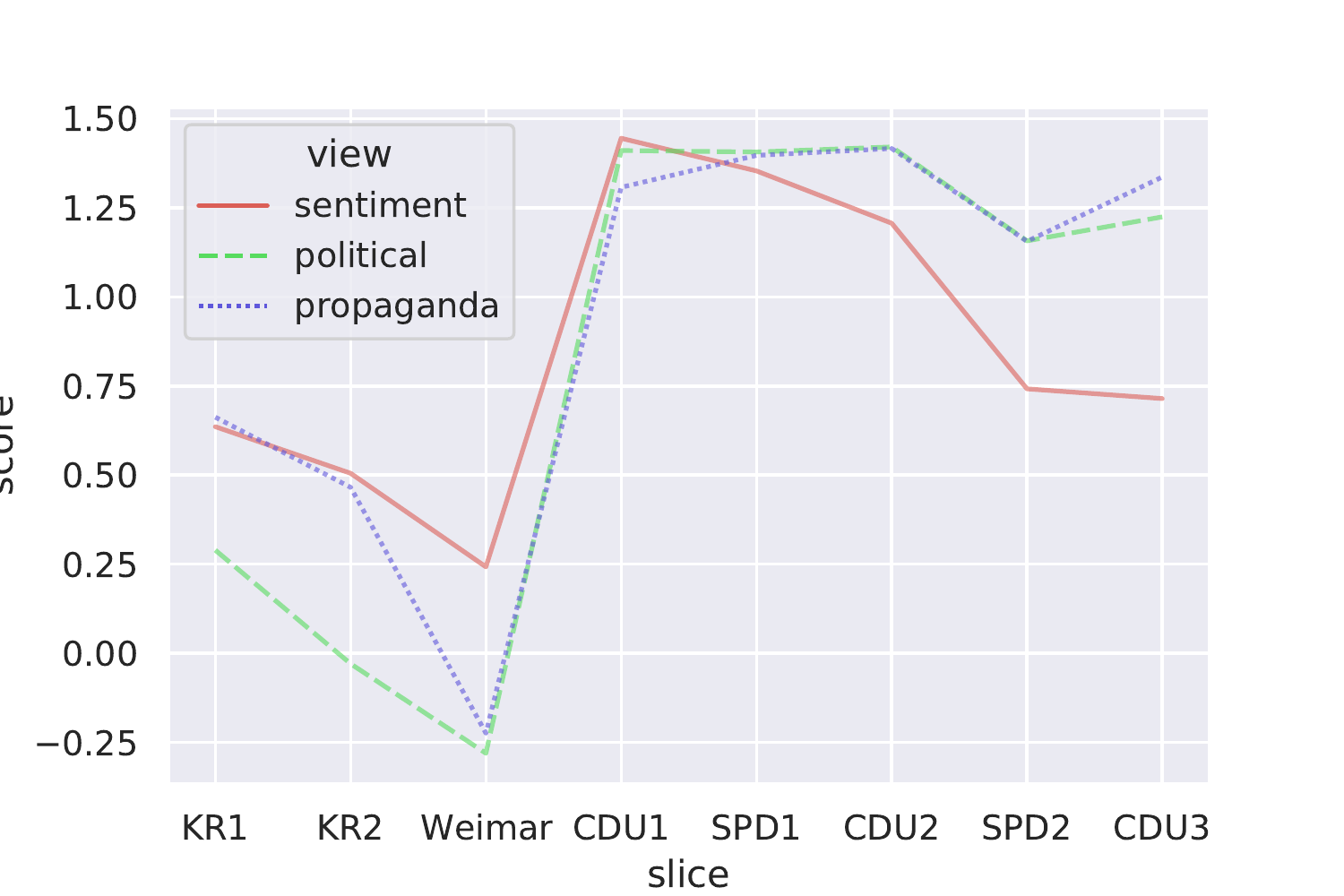}
         \caption{WEAT scores (anti-communism).}
         \label{fig:weat_anticommunism}
     \end{subfigure}
         \begin{subfigure}[t]{0.31\textwidth}
         \centering
         \includegraphics[width=1.0\linewidth,trim=0.0cm 0cm 1.5cm 0cm]{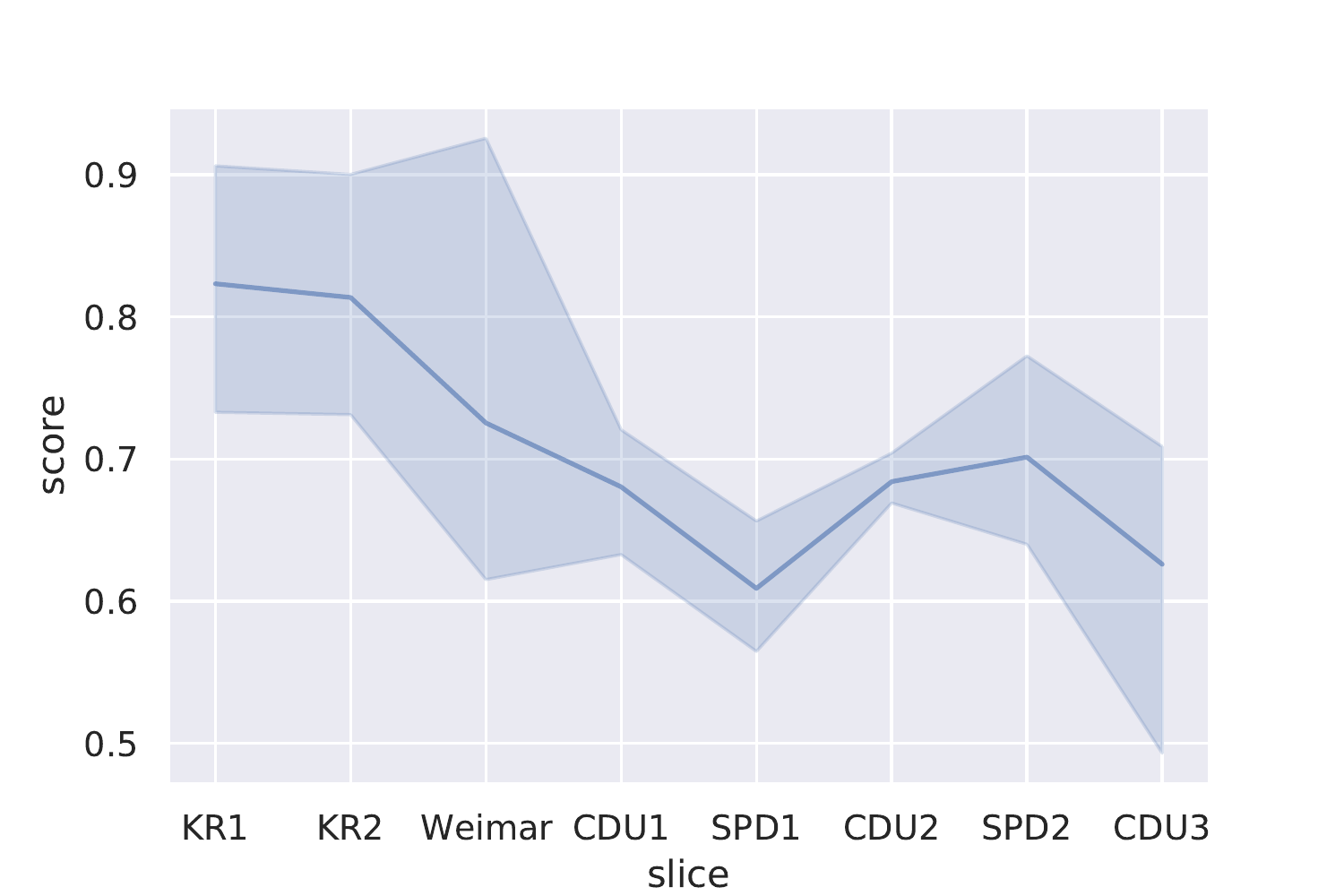}
         \caption{\footnotesize Aggregated ECT scores (anti-communism).}
         \label{fig:aggr_ect_anticommunism}
     \end{subfigure}
        \caption{Results of the embedding-based approach. WEAT: higher scores means lower bias; ECT: higher scores, lower bias.}
        \label{fig:embeddings}
\end{figure*}

\begin{figure*}[t!]
     \begin{subfigure}[h]{0.49\textwidth}
         \centering
         \includegraphics[width=1.0\linewidth,trim=0.0cm 0cm 0cm 0cm]{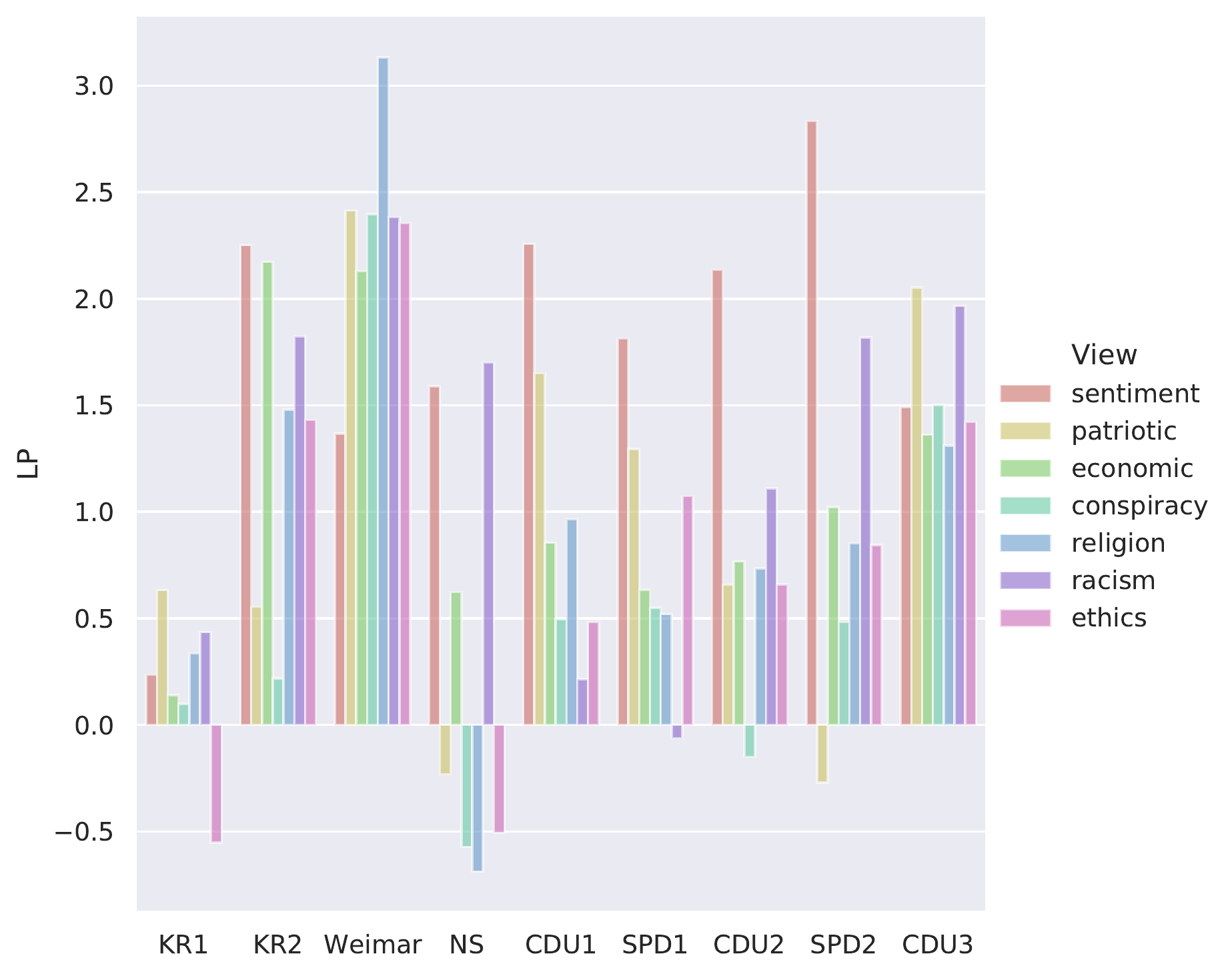}
         \caption{CBT scores (antisemitism).}
         \label{fig:hflp_antisemitism}
    \end{subfigure}
    \hfill
    \begin{subfigure}[h]{0.49\textwidth}
         \centering
         \includegraphics[width=1.0\linewidth,trim=0.0cm 0cm 0cm 0cm]{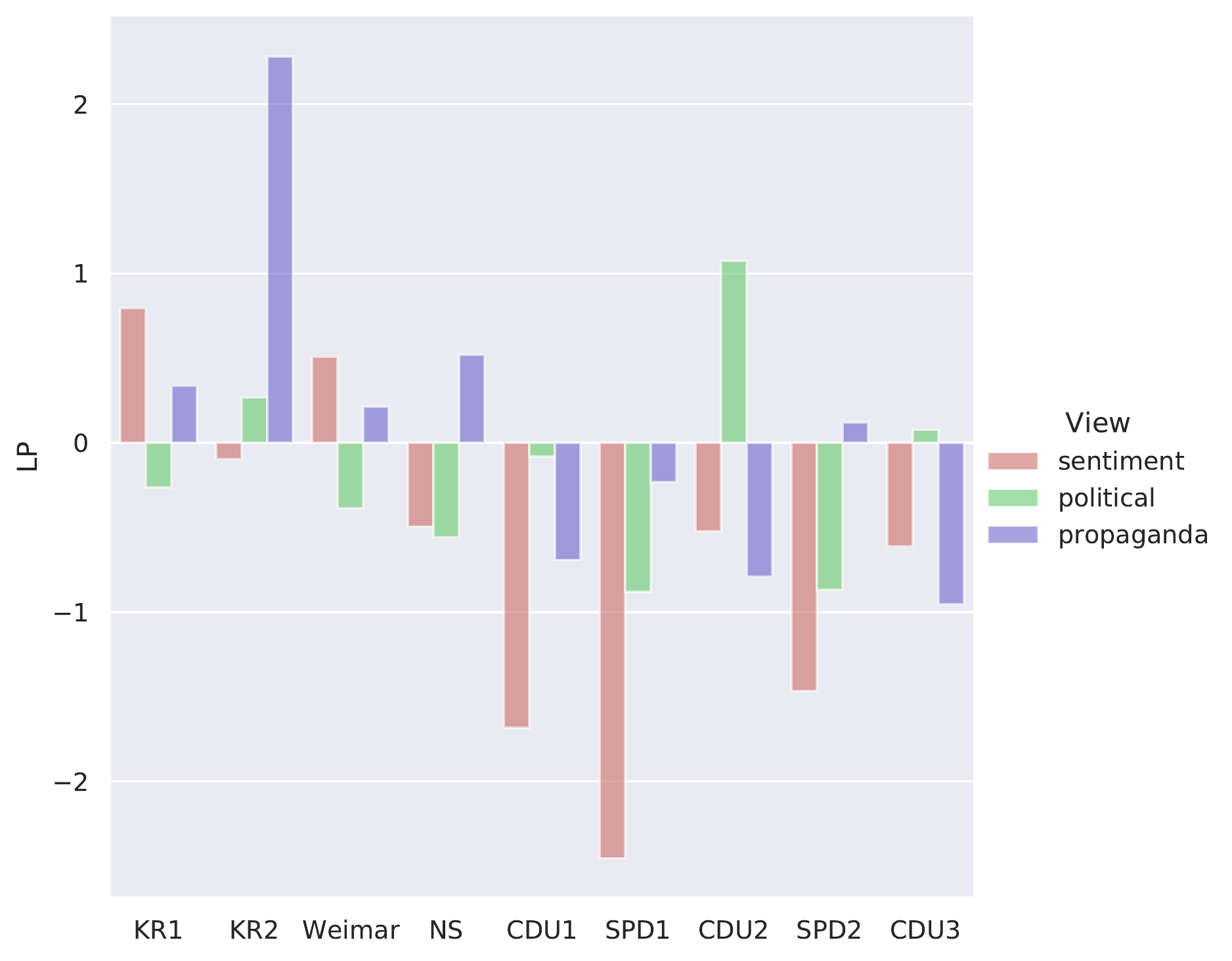}
         \caption{CBT scores (anti-communism).}
         \label{fig:hflp_anticommunism}
     \end{subfigure}
        \caption{Results for the graph-based approach.}
        \label{fig:graph}
\end{figure*}

The results for our embedding-based and the graph-based bias measurement approaches are depicted in Figures~\ref{fig:embeddings} and Figures~\ref{fig:graph}, respectively. In the following, we first focus on antisemitic bias and then discuss the results for the anti-communist bias.

\vspace{1em}
\noindent
\textbf{Antisemitic Bias.} Figure~\ref{fig:aggr_weat_antisemitism} shows the mean and $95\%$ confidence interval aggregated across the WEAT effect sizes for all bias views. 
By our specification, $T_1$ refers to the Christian group, $T_2$ to the Jewish group, $A_1$ to positively connotated words and $A_2$ to negatively connotated words. Thus, a large value of WEAT can be interpreted as antisemitic bias.
As it can be seen, antisemitism measured by WEAT continuously rises starting from KR1 (Kaiserreich 1, $1867-1890$) over KR2 (Kaiserreich 2, $1890-1918$) until it reaches its peak in WR ($1918-1942$).
Afterwards, except for a slight increase in CDU2 ($1982-1998$), the measured bias continuously drops and falls under the original level from KR1. Although there are slight differences, overall trends measured by ECT are similar (lower correlation indicates higher bias): the global minimum of the ECT mean can be found in the historical slice from the Weimar Republik ($1918-1942$), and the estimated bias from CDU2 ($1982-1998$) is smaller than the estimates obtained between KR1 and WR. The different bias views draw an interesting picture: for instance, while patriotic bias seems to drop, the bias measured towards attributes reflecting sentiment appears to stay consistent, even in the period of the Bundesrepublik. A reason for this is the fact that the discourse in the German parliament shifted towards discussing Germany's responsibility for the Holocaust and the need for remembrance, which is \emph{per se} not antisemitically biased, but carries a negative sentiment. Generally, the results obtained using the graph-based approach (CBT), shown in Figure \ref{fig:hflp_anticommunism}, are less significant but confirm the observations obtained using the embedding-based approach. 

\vspace{1em}
\noindent
\textbf{Anti-communist Bias.} The mean of the aggregated WEAT scores for anti-communist bias obtained with different views is starting from a local optimum in KR1 ($1867-1890$), and then decreasing until they peak in CDU1 ($1949-1969$, see Figure~\ref{fig:aggr_weat_anticommunism}). Interestingly, compared to this finding, the ECT scores depicted in Figure~\ref{fig:aggr_ect_anticommunism})
show a different picture, while the overall trend remains similar: generally, the scores seem to vary more heavily and, measured with ECT (all scores are significant at $\alpha < 0.05$), the bias seems to steadily rise until SPD1, decrease a bit and then increase again to CDU3 ($2005-2020$). Intuitively, though the results are not entirely conclusive, all of these peaks as well as the shared trend of the increased bias over the years make sense: (i) We speculate that the peak in KR1 reflects the prominence of anti-socialist laws from that period; (ii) After the end of the monarchy in Germany, the communists formally founded the communist party Kommunistische Partei Deutschlands (KPD) and the SPD continued to establish itself as a moderate-leftist party, entering into
the Weimarer Coalition. Until the $1932$ election  (6th legislative period in WR), the SPD remained the strongest political force in
the Reichstag. Generally, the SPD gained
a reputation for pursuing compromises between the different ideological factions while the KPD strongly advocated for abolishing the parliamentary system and even constituted an anti-parliamentary alliance with the NSDAP in the
late WR years ~\cite{halder2003innenpolitik}. In a qualitative analysis, we found evidence indicating the moderate left’s desire to distance
themselves from extremist communism. For instance, they denounce the bad reflection of the communist tactics on honest desire to improve conditions for the working class; 
(ii) after World War II (CDU1), the establishment of the Iron Curtain led to
a clash of Western and Eastern ideologies~\cite{creuzberger2014antikommunismus}. In our study, we only include protocols from the 
Bundesrepublik, 
which reflect just the Western perspective.
(iii) The reason for the peak in CDU3 may be attributed to the major parties’
general stance against populist and anti-democratic forces, which have been gaining
momentum in recent years~\cite{jun2020parteien}. The different views 
indicate that, initially (KR1), the 
bias with respect to the sentiment attribute sets is higher, while later on, and especially in the last periods, it gets surpassed by the bias with respect to politics and propaganda terms. 

\vspace{1em}
\noindent
\textbf{Consistency across Measures.} As already noticed by Lauscher et al.~\cite{lauscher2020general}, different bias measurement approaches capture different aspects of bias in the distributional space and therefore sometimes deviate in terms of the amount of bias they reveal. In line with their findings, we note some inconsistencies: for instance, the amount of bias measured with CBT is generally lower than the one estimated with WEAT. This is not surprising, since WEAT has been shown to overestimate bias~\cite{ethayarajh-etal-2019-understanding}. We therefore acknowledge the importance of running several measures in parallel in order to capture the most pervasive trends.

\section{Related Work}

\vspace{1em}
\noindent
\textbf{Measuring Bias in Semantic Spaces.} Bolukbasi et al.~\cite{bolukbasi2016man} were the first to demonstrate that word embeddings induced from human-created corpora encode stereotypical human biases. More specifically, they showed that many distributional vector spaces allow for algebraically building biased analogies, such as the famous example \emph{``man is to computer programmer as woman is to homemaker''}, an example of a sexist analogy. Following up on this, Caliskan et al.~\cite{caliskan2017semantics} presented WEAT, inspired from the Implicit Association Test ~\cite{nosek2002harvesting} from psychology, which measures biases in terms of associative difference in similarity between term sets. The WEAT bias specifications were later translated to several languages, e.g., by Chaloner and Maldonado~\cite{chaloner-maldonado-2019-measuring} and Lauscher et al.~\cite{lauscher-glavas-2019-consistently,lauscher-etal-2020-araweat}. We use the German sentiment attribute sets from the latter. A similar group of authors assembled a larger framework of bias evaluation measures~\cite{lauscher2020general} and proposed the notion of implicit and explicit bias specifications. This framework includes, for instance, the Embedding Coherence Test~\cite{dev2019attenuating} and the Implicit Bias Test \cite{gonen2019lipstick}. We adopt their notion of bias. Follow up research also proposed evaluation methods and resources for contextualized embedding methods~\cite[\emph{inter alia}]{zhao-etal-2019-gender,kurita-etal-2019-measuring,basta-etal-2019-evaluating}.

\vspace{1em}
\noindent
\textbf{Natural Language Processing for the Analysis of Historical Corpora.} The studies closest to ours were presented by Garg et al.~\cite{Garg.2018} and Tripodi et al.~\cite{Tripodi.}. Similar to us, Garg et al.\ quantify bias in historical corpora employing word vector spaces as proxies. However, they analyze gender and ethnic bias in U.S.-centric corpora, while we specifically focus on antisemitic and anti-communist bias in German parliamentary data, which allows us to capture ideological trends.
Tripodi et al. focus, similar to us, on antisemitic bias, but they analyze French books and periodicals issues. We additionally propose a graph-based bias evaluation, which allows us to deal with smaller historical slices. Using word embeddings for studying change in historical corpora was 
introduced by Hamilton et al.~\cite{hamilton-etal-2016-diachronic} and Eger and Mehler \cite{eger-mehler-2016-linearity}: however, while their contribution is on determining statistical laws of semantic change in diachronic word embeddings, we focus here instead on whether changes in bias measurements from dense and sparse embeddings align against known historical changes. Despite their enormous popularity, there exist alternatives to word embeddings as methodology to analyze historical corpora, including topic modeling \cite{yang-etal-2011-topic,blei12} and event extraction \cite{rovera21}.

\section{Conclusion}
In this work, we have investigated how to employ the notion of bias in distributional word vector spaces for understanding ideological shifts in a large historical corpus of German parliamentary debates. The results obtained with the various available measures sometimes exhibit different amounts of bias. However, many shared trends can be found, which can, in turn, be attributed to events related to ideological shifts within the German history, e.g., the establishment of the iron curtain or peaks in antisemitism 
leading up to 
the NS time.
Along with our study, we release \corpus hoping to fuel further research on computational understanding of German parliamentary proceedings.

Using the notion of bias, we are able to capture commonly accepted historical trends. While the embedding-based method has the advantage that a large range of bias evaluation measures for estimating the extend of stereotyping present in text representation models is available, it has the drawback in that the sizes of the slices need to be sufficiently large to induce reliable embeddings. In these cases, we proposed to `fall-back' to more traditional meaning representations, i.e., co-occurrence graphs, and leverage label propagation algorithms to infer bias from sparse co-occurrence counts.

Our analysis shows that main ideological shifts can be read from historical corpora through the quantification of bias of semantic spaces induced from them. Our method has the potential to enable a bias-centric exploration of textual collections from experts and thus calls for the future integration of our quantitative analysis with qualitative ones in the manner of \emph{distant reading} from historians. We also plan to tackle other languages than German, for a contrastive evaluation of multilingual cross-temporal bias, in future work.

\section*{Acknowledgments}

We thank the reviewers for insightful comments. We thank Niklas Friedrich for the independent reimplementation of the approaches and reproduction of the empirical results.

\bibliographystyle{IEEEtranS}
\bibliography{IEEEabrv,main}
\end{document}